\documentclass[journal]{IEEEtran}

\usepackage{amsmath,amsfonts}
\usepackage{amssymb}

\usepackage{graphicx}
\usepackage[caption=false,font=footnotesize,labelfont=sf,textfont=sf]{subfig}

\usepackage{booktabs}
\usepackage{multirow}
\usepackage{array}

\usepackage{algorithmic}
\usepackage{algorithm}

\usepackage{textcomp}
\usepackage{stfloats}
\usepackage{url}
\usepackage{verbatim}

\usepackage{cite}

\usepackage{hyperref}
\hypersetup{
    colorlinks=true,
    linkcolor=black,
    citecolor=green,
    urlcolor=black
}
\usepackage{orcidlink} 
\makeatletter
\newcommand{\notetopractitioners}{%
  \vspace{0.67ex}%
  \noindent\@IEEEabskeysecsize\bfseries\textit{Note to Practitioners}---}
\makeatother

\begin{document}

\title{ICD-Net: Inertial Covariance Displacement Network for Drone Visual-Inertial SLAM}

\author{Tali~Orlev~Shapira~\orcidlink{0009-0001-4385-9638},~
\thanks{T. Orlev Shapira is with the Hatter Department of Marine Technologies,
Charney School of Marine Sciences, University of Haifa, Israel
(email: taliorlev11@gmail.com).}%
Itzik~Klein~\orcidlink{0000-0001-7846-0654}%
\thanks{I. Klein is with the Hatter Department of Marine Technologies,
Charney School of Marine Sciences, University of Haifa, Israel
(email: kitzik@univ.haifa.ac.il).}%
}

\maketitle
\begin{abstract}
Visual-inertial SLAM systems often exhibit suboptimal performance due to multiple confounding factors including imperfect sensor calibration, noisy measurements, rapid motion dynamics, low illumination, and the inherent limitations of traditional inertial navigation integration methods. These issues are particularly problematic in drone applications where robust and accurate state estimation is critical for safe autonomous operation. In this work, we present ICD-Net, a novel framework that enhances visual-inertial SLAM performance by learning to process raw inertial measurements and generating displacement estimates with associated uncertainty quantification. Rather than relying on analytical inertial sensor models that struggle with real-world sensor imperfections, our method directly extracts displacement maps from sensor data while simultaneously predicting measurement covariances that reflect estimation confidence. We integrate ICD-Net outputs as additional residual constraints into the VINS-Fusion optimization framework (a visual-inertial SLAM approach), where the predicted uncertainties appropriately weight the neural network contributions relative to traditional visual and inertial terms. The learned displacement constraints provide complementary information that compensates for various sources of errors in the SLAM pipeline. Our approach can be used under both normal operating conditions and in situations of camera inconsistency or visual degradation. Experimental evaluation of challenging high-speed drone sequences demonstrated that our approach significantly improved trajectory estimation accuracy compared to standard VINS-Fusion, with more than 38\% improvement in mean APE and uncertainty estimates proving crucial for maintaining system robustness. Our method shows that neural network enhancement can effectively address multiple sources of SLAM degradation while maintaining real-time performance requirements.
\end{abstract}

\section{Introduction}
\IEEEPARstart{A}{utonomous} navigation systems for unmanned aerial vehicles (UAVs) rely heavily on accurate state estimation to ensure safe and reliable operation. Visual-inertial simultaneous localization and mapping (SLAM) approaches have emerged as a dominant paradigm, combining complementary information from cameras and inertial measurement units (IMUs) to achieve robust pose estimation across diverse environments~\cite{huang2019visual,servieres2021visual,wang2024review,10287884}. Recent comprehensive surveys have highlighted both the evolution and ongoing challenges in visual SLAM systems~\cite{alzubaidi2022comprehensive,frontiers2024review}. High-fidelity simulation platforms for integrated aerial systems have been developed to facilitate research in control, perception, and learning for these challenging scenarios~\cite{10942427}. However, achieving optimal SLAM performance in real-world conditions remains challenging due to multiple confounding factors including sensor noise, calibration imperfections, rapid motion dynamics, and the fundamental limitations of sensor fusion algorithms. Modern SLAM systems face particular difficulties when operating with low-cost sensors in high-dynamic scenarios. IMU integration, which relies on numerical integration of specific force and angular velocity measurements, is inherently susceptible to noise amplification and drift accumulation~\cite{farrell2008aided,titterton2004strapdown}. These problems are compounded by imperfect sensor calibration, time synchronization errors, and the complex noise characteristics of consumer-grade inertial sensors~\cite{schmidt2024visual,reitbauer2023liwo}. Modern approaches have attempted to address these limitations through multi-sensor fusion strategies~\cite{improving2024slam}, enhanced computational frameworks~\cite{tsintotas2022robust}, and deep-learning approaches~\cite{solodar2024vio}. In drone applications, where platforms can experience multi-g accelerations and rapid orientation changes, these limitations often manifest as tracking failures and divergent state estimates that compromise system reliability. The severity of these challenges becomes apparent when evaluating SLAM systems on demanding datasets such as UZH-FPV~\cite{delmerico2019uzh}, where high-speed drone flights with aggressive maneuvers expose the weaknesses of traditional sensor fusion approaches. Even state-of-the-art systems like VINS-Fusion~\cite{qin2019general}, ORB-SLAM3~\cite{campos2021orb}, and others struggle to maintain consistent performance under such extreme conditions, highlighting the need for more robust integration methods that can handle real-world sensor imperfections and dynamic motion patterns.
Recent advances in deep learning~\cite{chen2024deep,cohen2024inertial} have demonstrated the potential for neural networks to learn complex sensor models and improve state estimation in robotics applications~\cite{sl2024slam,rover2024slam,cohen2025adaptive}. Learning-based approaches for IMU processing and calibration have shown promise in addressing sensor imperfections~\cite{calib2022net,stolero2026rapid}, while hybrid visual SLAM frameworks have demonstrated improved performance in challenging environments~\cite{wang2024switching}. Rather than relying on analytical models of sensor behavior, learning-based approaches can adapt to specific sensor characteristics and operational conditions. Previous works that have attempted neural network-based IMU processing include TLIO~\cite{liu2020tlio}, RNIN-VIO~\cite{chen2021rnin}, and IONet~\cite{chen2018ionet}. However, these approaches are primarily based on pedestrian and smartphone data, focusing on human motion patterns that differ significantly from the high-dynamic maneuvers characteristic of drone flight~\cite{klein2025pedestrian}. Most other existing work focuses on improving visual processing or end-to-end learning, with limited attention to enhancing IMU integration for UAV applications while maintaining the proven optimization frameworks of established SLAM systems.\\
Standard visual-inertial SLAM systems often struggle with imperfect inertial sensor calibration, noisy measurements, and environmental conditions that influence the inertial readings. In addition, in poor lightning conditions and rapid motion dynamics, degradation in the vision part performance is evident. To address these limitations, we propose ICD-Net, a novel neural network approach that directly processes raw inertial measurements to estimate the displacement vector with its associated uncertainty quantification. Unlike analytical models that struggle with real-world sensor imperfections, ICD-Net learns to extract displacement information while simultaneously predicting measurement covariances that reflect estimation confidence. Our method complements standard inertial preintegration by providing learned displacement constraints that capture complex sensor dynamics and noise patterns beyond what conventional models can represent. The ICD-Net outputs are integrated as additional residual constraints into the VINS-Fusion optimization framework, where the predicted uncertainties appropriately weight the neural network contributions relative to standard visual and inertial terms.\\
Our key contributions are: 
\begin{enumerate}
\item \textbf{ICD-Net:} A two-heads neural network architecture that learns displacement estimation from inertial sensor data with uncertainty quantification.
\item \textbf{Enhanced VINS-Fusion:} A graph optimization integration strategy that incorporates learned inertial sensor constraints into existing VINS-Fusion optimization frameworks. 
\item \textbf{Robustness To Camera Inconsistency:} As our approach leverages inertial constraints to maintain performance during visual data loss, it offers substantial robustness to camera inconsistencies and camera blackouts. The inertial constraints act as a reliable backup, allowing the system to operate smoothly and provide a continuous, accurate state estimate. This mitigates inertial drift and makes the overall system more accurate and robust.
\end{enumerate}
\par\noindent To validate our approach, we employed the UZH-FPV dataset, which contains challenging high-speed drone sequences. Our results show that the proposed method overcomes the shortcomings of VINS-Fusion during aggressive flight maneuvers, while preserving real-time computational efficiency. This is particularly vital for high-speed drones, where visual data degrades due to motion blur, making reliable IMU integration crucial for stable pose estimation.\\
The rest of the paper is organized as follows: Section~\ref{sec:prop_app} presents the proposed approach - the ICD-Net and the VINS-Fusion integration, Section~\ref{sec:results} provides dataset information and analysis of the results, and Section~\ref{sec:conc} concludes this study.
\section{Proposed Approach}
\label{sec:prop_app}
\subsection{Motivation}
\noindent Visual-inertial SLAM systems face fundamental challenges when deployed on UAV platforms. The six-degree-of-freedom motion characteristic of aerial vehicles-unconstrained rotation and translation in 3D space-creates conditions that systematically stress the sensing modalities. Unlike ground robots with planar motion constraints, UAVs exhibit unpredictable and highly dynamic trajectories including rapid rotations, sudden directional changes, and varying velocities. These motion patterns degrade camera measurements through motion blur and feature tracking failures while amplifying noise in the low-cost IMUs typically used on drones. During high-speed flights, this challenge becomes particularly acute as camera performance degrades significantly, while the system must rely heavily on IMU measurements for state estimation.\\
Existing approaches struggle to handle these varying conditions due to rapid inertial drift and their reliance on fixed noise models. Conventional pre-integration methods assume static noise characteristics, failing to account for how measurement uncertainty varies across different flight conditions-stable hovering versus aggressive maneuvers produce fundamentally different error profiles. This leads to suboptimal sensor weighting during optimization, where the system treats all IMU measurements with the same assumed noise level regardless of the actual motion context that produced them.\\
Our approach addresses these challenges by using neural networks to learn the relationship between raw IMU measurements
and the platform displacement vector while simultaneously
predicting measurement uncertainties. This data-driven method
can adapt to specific sensor noise characteristics and motion
patterns, providing uncertainty estimates that reflect actual
measurement reliability rather than fixed theoretical models. By enabling motion-aware uncertainty quantification, our method allows the SLAM optimization framework to appropriately weight IMU constraints based on their current reliability under both normal operating conditions and situations of camera inconsistency or visual degradation.
\subsection{Method Overview}
\begin{figure}[!t]
    \centering
    \includegraphics[width=0.7\columnwidth]{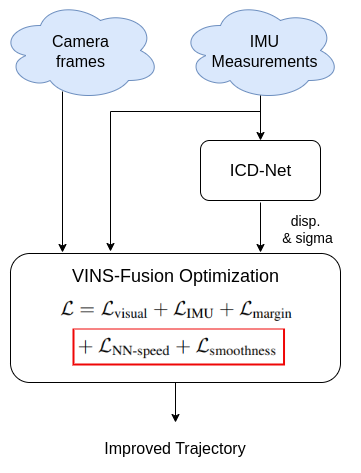}
    \caption{Enhanced VINS-Fusion framework with our proposed modifications. We introduce ICD-Net for learning displacement and uncertainty estimates from inertial data, and augment the VINS-Fusion optimization with additional neural network-based loss terms (highlighted in red): $\mathcal{L}_{\text{NN-speed}}$ and $\mathcal{L}_{\text{smoothness}}$.}
    \label{fig:big_pic}
\end{figure}

\noindent Figure \ref{fig:big_pic} illustrates the complete pipeline of our proposed system. The process begins with ICD-Net, a neural network that receives raw IMU measurements and outputs both displacement predictions and their associated covariance estimates. We first train this network on the training dataset to learn the relationship between IMU measurements and platform motion across various flight conditions. After training, the system operates in real-time by incorporating ICD-Net's outputs of the test dataset into the VINS-Fusion optimization framework. For each test sequence, ICD-Net processes the incoming IMU measurements and feeds its predictions to the VINS-Fusion optimization.\\
We extend the VINS-Fusion optimization cost function with two additional factors, highlighted in red in the figure: $\mathcal{L}_{\text{NN-speed}}$ and $\mathcal{L}_{\text{smoothness}}$. The speed constraint, which is weighted by the predicted covariances, encourages the network to produce displacement estimates that align with expected motion dynamics. The smoothness term prevents abrupt changes in uncertainty predictions that could destabilize the optimization. In the following subsections, we elaborate on each component of our proposed approach.
\subsection{Neural Network Architecture}
\begin{figure}[!t]
    \centering
    \includegraphics[width=\columnwidth]{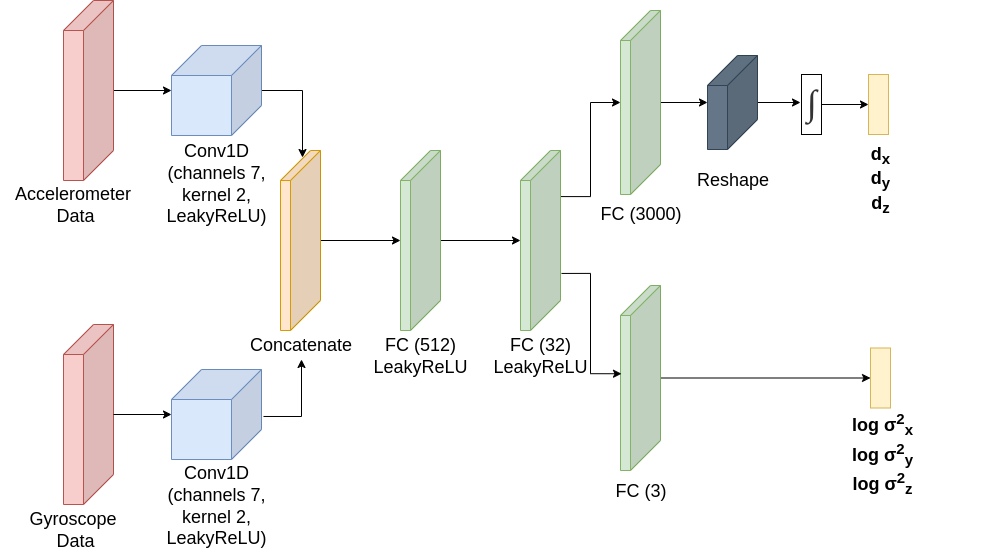}
    \caption{The ICD-Net architecture uses two heads to process the inertial readings and output the displacement and covariance estimates.}
    \label{fig:nn}
\end{figure}

\noindent Our proposed neural network, ICD-Net, illustrated in Figure \ref{fig:nn}, processes raw IMU measurements to estimate platform displacement and predict associated measurement uncertainty. \\
Let $\mathbf{f}_t \in \mathbb{R}^3$ and $\boldsymbol{\omega}_t \in \mathbb{R}^3$ denote the measured specific force vector and angular velocity vector at time $t$, respectively. For a window of $T$ consecutive measurements, the input sequences are:
\begin{align}
\mathbf{F} &= [\mathbf{f}_1, \mathbf{f}_2, \ldots, \mathbf{f}_T] \in \mathbb{R}^{3 \times T} \\
\boldsymbol{\Omega} &= [\boldsymbol{\omega}_1, \boldsymbol{\omega}_2, \ldots, \boldsymbol{\omega}_T] \in \mathbb{R}^{3 \times T}
\end{align}
The network learns a mapping $f_\theta: \mathbb{R}^{3 \times T} \times \mathbb{R}^{3 \times T} \rightarrow \mathbb{R}^3 \times \mathbb{R}^3$ that produces displacement and uncertainty estimates for the entire window:
\begin{equation}
(\mathbf{d}, \boldsymbol{\sigma}^2) = f_\theta(\mathbf{F}, \boldsymbol{\Omega})
\end{equation}
where $\mathbf{d} = [d_x, d_y, d_z]^T \in \mathbb{R}^3$ is the predicted displacement vector and $\boldsymbol{\sigma}^2 = [\sigma_x^2, \sigma_y^2, \sigma_z^2]^T \in \mathbb{R}^3$ is the corresponding predicted variance vector, reflecting the uncertainty of the displacement vector.\\
The key challenge is learning this mapping such that the predicted displacement $\mathbf{d}$ accurately reflects the true motion while the uncertainty estimates $\boldsymbol{\sigma}^2$ provide reliable confidence measures for integration into the VINS-Fusion optimization framework.
\subsubsection{Network Structure}
The network takes as input one-second windows of IMU data, consisting of $T$ samples. Accelerometer $\mathbf{f} \in \mathbb{R}^{3 \times T}$ and gyroscope $\mathbf{\omega} \in \mathbb{R}^{3 \times T}$ measurements are fed as separate input channels. The architecture consists of three main components:\\
\textbf{Feature Extraction:} Separate 1D convolutional layers process accelerometer and gyroscope data independently to capture temporal patterns in the sensor measurements:
\begin{align}
\mathbf{c}_{f} &= \text{Conv1D}(\mathbf{f}) \\
\mathbf{c}_{\omega} &= \text{Conv1D}(\mathbf{\omega})
\end{align}
Each convolutional branch uses kernel size $2$, stride $2$, and produces $7$ feature channels, followed by LeakyReLU~\cite{maas2013rectifier} activation and dropout for regularization. The raw IMU data is also preserved and concatenated with the extracted features to maintain fine-grained sensor information.\\
\textbf{Feature Fusion:} The flattened convolutional features and raw sensor data are concatenated and processed through fully-connected layers with layer normalization:
\begin{equation}
\mathbf{h} = \text{FC}_2(\text{FC}_1([\mathbf{c}_{f}; \mathbf{c}_{\omega}; \mathbf{f}; \mathbf{\omega}]))
\end{equation}
\textbf{Dual-Head Output:} The fused features are fed into two separate decoder heads:
\begin{align}
\mathbf{v}_{pred} &= \text{VelocityDecoder}(\mathbf{h}) \in \mathbb{R}^{\frac{T}{2} \times 3} \\
\log \boldsymbol{\sigma}^2 &= \text{LogVarDecoder}(\mathbf{h}) \in \mathbb{R}^{3}
\end{align}
The velocity decoder produces $T/2$ velocity predictions, which are integrated to obtain the final displacement over the one-second window:
\begin{equation}
\mathbf{d}_{pred} = \sum_{i=1}^{T/2} \mathbf{v}_{pred,i} \cdot \Delta t
\end{equation}
where $\mathbf{v}_{pred,i}$ is the predicted velocity vector at time $i$, $\mathbf{d}_{pred}$ is the predicted displacement vector, and $\Delta t$ is the output sampling interval.\\
The log-variance decoder produces window-level uncertainty estimates $\log \sigma^2_x$, $\log \sigma^2_y$, $\log \sigma^2_z$ for each spatial axis. The network outputs log-variance to naturally enforce positivity constraints and to provide better gradient stability during training. The final variances are obtained via exponentiation.\\
\subsubsection{Loss Function}
We employ a physics-informed loss function that combines displacement accuracy, motion smoothness, and uncertainty calibration:
\begin{equation}
\label{eq:total_loss}
\mathcal{L}_{total} = \alpha \mathcal{L}_{base} + \beta \mathcal{L}_{reg} + \gamma \mathcal{L}_{NLL}
\end{equation}
where $\alpha$, $\beta$, and $\gamma$ are weighting coefficients that balance the three loss components. Next, we define each of the loss terms in Eq.(\ref{eq:total_loss}).\\
\textbf{Base Loss:} Combines L1 displacement loss with velocity smoothness regularization:
\begin{align}
\mathcal{L}_{base} &= \|\mathbf{d}_{pred} - \mathbf{d}_{GT}\|_1 \nonumber \\
&\quad + \lambda_{smooth} \sum_{t=2}^{500} \left\|\frac{\mathbf{d}_{pred}[t] - \mathbf{d}_{pred}[t-1]}{\Delta t}\right\|_2^2
\end{align}
where $\mathbf{d}_{GT} \in \mathbb{R}^3$ is the ground truth displacement, $\lambda_{smooth}$ is the smoothness weight, and $\Delta t = 0.002$ seconds.\\
\textbf{Negative Log-Likelihood Loss:} A critical aspect of our approach is training the network to produce meaningful uncertainty estimates $\boldsymbol{\sigma}^2$. Since ground truth uncertainty values are not available in the training data, we employ a negative log-likelihood (NLL) loss term that treats the predicted variances as parameters of a probabilistic model.
Specifically, we model the prediction error as a multivariate Gaussian distribution:
\begin{equation}
p(\mathbf{d}_{gt} | \mathbf{d}_{pred}, \boldsymbol{\sigma}^2) = \prod_{k=1}^{3} \frac{1}{\sqrt{2\pi\sigma_k^2}} \exp\left(-\frac{(d_{gt,k} - d_{pred,k})^2}{2\sigma_k^2}\right)
\end{equation}
The negative log-likelihood loss becomes:
\begin{equation}
\begin{split}
\mathcal{L}_{NLL} &= -\log p(\mathbf{d}_{gt} | \mathbf{d}_{pred}, \boldsymbol{\sigma}^2) \\
&= \frac{1}{2}\sum_{k=1}^{3}\left[\log\sigma_k^2 + \frac{(d_{gt,k} - d_{pred,k})^2}{\sigma_k^2} + \log(2\pi)\right]
\end{split}
\end{equation}
where $k \in \{x,y,z\}$ indexes the spatial dimensions.
This formulation encourages the network to predict small variances $\sigma_k^2$ when prediction errors are small, and large variances when errors are large, effectively learning to calibrate its own confidence. The NLL loss ensures that the variance outputs represent genuine uncertainty rather than arbitrary values, making them suitable for weighting constraints in the SLAM optimization process.\\
\textbf{Regularization Loss:} Prevents extreme uncertainty estimates through per-axis L2 regularization:
\begin{equation}
\mathcal{L}_{reg} = \sum_{k} \lambda_{reg,k} (\log \sigma_k^2)^2
\end{equation}
where $\lambda_{reg,k}$ is the regularization weight for axis $k$ and $\sigma_k^2$ is the predicted variance for the corresponding spatial axis.\\
The log-variance outputs are clamped to $[\log \sigma_{min}^2, \log \sigma_{max}^2]$ to ensure numerical stability. This multi-term loss function ensures that the network learns both accurate displacement predictions and reliable uncertainty estimates, which are crucial for proper weighting in the VINS-Fusion optimization framework.\\
\subsubsection{Optimization Strategy}
The training process for our network utilizes a robust optimization strategy that combines the AdamW optimizer~\cite{loshchilov2017decoupled} with a ReduceLROnPlateau scheduler~\cite{pytorch_reducelr}. This two-pronged approach ensures both stable convergence and efficient fine-tuning of the model parameters. We employed the AdamW optimizer, a modern variant of the widely used Adam algorithm, which decouples the weight decay from the gradient update to enhance training stability and improve generalization. The optimizer was configured with a learning rate of $0.002$ and a weight decay of $1 \times 10^{-2}$. To dynamically adjust the learning rate and prevent overfitting, we utilized a ReduceLROnPlateau scheduler. This scheduler monitors a performance metric (e.g., validation loss) and automatically reduces the learning rate by a factor of $0.5$ if the metric plateaus for a patience of 10 epochs. This adaptive mechanism allows for aggressive learning early in training and a more refined search for the optimal solution as the model converges.\\
The network was trained over a total of 300 epochs, with a staged approach to the loss function hyperparameters. The training was conducted in two distinct phases to optimize the network's performance.
During the initial 100 epochs, which served as a warm-up phase, the following hyperparameters were employed for the loss function (\ref{eq:total_loss}):
\begin{align}
\alpha = 1, \quad \lambda_{\text{smooth}} = 5 \times 10^{-5}, \quad \beta = 0, \quad \gamma = 0
\end{align}
For the subsequent 200 epochs, the training continued with an updated set of hyperparameters, specifically introducing non-zero values for $\beta$ and $\gamma$ to modify the loss behavior:
\begin{align}
\alpha = 1, \quad \lambda_{\text{smooth}} = 5 \times 10^{-5}, \quad \beta = 0.1, \quad \gamma = 8
\end{align}
\subsection{Integration with VINS-Fusion}
\noindent We adopt VINS-Fusion as our backbone framework primarily due to its graph optimization approach, which provides certain advantages for integrating neural network predictions. We note that similar integration could be achieved with filter-based methods such as OpenVINS~\cite{geneva2020openvins} and ROVIO~\cite{bloesch2015robust}.\\
VINS-Fusion's sliding window bundle adjustment optimizes over a history of poses and landmarks, allowing the system to refine past state estimates when new neural network predictions become available. Its factor graph formulation allows us to add the neural network displacement residuals as additional constraints in the Ceres Solver optimization~\cite{ceres-solver}, where automatic differentiation handles all derivative computations. This approach naturally accommodates the uncertainty estimates from our neural network by directly incorporating predicted covariances as information matrices for the neural network residuals. The optimization framework automatically balances these learned constraints against standard IMU pre-integration factors and visual reprojection errors based on their relative uncertainties.\\
VINS-Fusion has demonstrated robust performance on challenging datasets and maintains a real-time operation suitable for drone applications. The system’s ability to jointly optimize visual and inertial measurements over multiple time steps provides significant resilience to aggressive maneuvers.\\
Our integration approach adds two types of residual terms to the VINS-Fusion optimization problem. The complete cost function becomes:
\begin{equation}
\mathcal{L} = \mathcal{L}_{\text{visual}} + \mathcal{L}_{\text{IMU}} + \mathcal{L}_{\text{margin}} + \mathcal{L}_{\text{NN-speed}} + \mathcal{L}_{\text{smoothness}}
\end{equation}
where $\mathcal{L}_{\text{visual}}$ and $\mathcal{L}_{\text{IMU}}$ are the original visual reprojection and IMU pre-integration terms, respectively, $\mathcal{L}_{\text{margin}}$ represents marginalization priors from the sliding window optimization, and $\mathcal{L}_{\text{NN-speed}}$ and $\mathcal{L}_{\text{smoothness}}$ are our neural network-derived constraints defined below.\\
\textbf{Neural Network Velocity Constraint:}
Our neural network processes IMU measurements acquired between consecutive keyframes $i$ and $j$ to predict displacement $\mathbf{d}_{\text{NN}}^{i \rightarrow j}$ and associated covariance $\boldsymbol{\Sigma}_{\text{NN}}^{i}$. We derive a velocity estimate from this prediction:
\begin{equation}
\mathbf{v}_{\text{NN}}^{i} = \frac{\mathbf{d}_{\text{NN}}^{i \rightarrow j}}{\Delta t}
\end{equation}
where $\Delta t$ is the time interval between keyframes. This network-derived velocity serves as an additional constraint in the VINS-Fusion optimization, encouraging its velocity estimate $\mathbf{v}_{\text{VINS}}^{i}$ to remain consistent with the neural network prediction. We incorporate this constraint through the following residual term:
\begin{equation}
\mathcal{L}_{\text{NN-vel}} = \sum_{i} \left\|\mathbf{v}_{\text{VINS}}^{i} - \mathbf{v}_{\text{NN}}^{i}\right\|_{\boldsymbol{\Omega}_{\text{vel}}^{i}}^2
\end{equation}
where $\boldsymbol{\Omega}_{\text{vel}}^{i}$ is the information matrix derived from the neural network's predicted covariance. The information matrix is the inverse of the covariance matrix, which ensures that measurements with lower uncertainty (higher confidence) receive higher weight in the optimization, and vice versa. Since velocity is obtained by differentiating position, we model the velocity uncertainty as:
\begin{equation}
\boldsymbol{\Omega}_{\text{vel}}^{i} = \text{diag}\left(\frac{1}{\sigma_{\text{NN},x}^2}, \frac{1}{\sigma_{\text{NN},y}^2}, \frac{1}{\sigma_{\text{NN},z}^2}\right)
\end{equation}
where $\sigma_{\text{NN},k}^2$ are the diagonal elements of the neural network's predicted covariance matrix for each axis $k \in \{x,y,z\}$.\\
Note that while we directly constrain the velocity estimates in VINS, this optimization naturally affects the position estimates since positions are integrated from velocities over the sliding window, and the bundle adjustment jointly optimizes all state variables including poses, velocities, and biases.\\
\textbf{Velocity Smoothness Constraint:}
To regularize the velocity estimates and prevent unrealistic acceleration spikes, we add a smoothness constraint that penalizes large changes in velocity between consecutive frames:
\begin{equation}
\mathcal{L}_{\text{smoothness}} = \sum_{i} \left\|\frac{\mathbf{v}_{\text{VINS}}^{i+1} - \mathbf{v}_{\text{VINS}}^{i}}{\Delta t}\right\|_{\boldsymbol{\Omega}_{\text{accel}}}^2
\end{equation}
This term constrains the implied acceleration $\mathbf{a}^{i} = (\mathbf{v}^{i+1} - \mathbf{v}^{i})/\Delta t$ to remain within reasonable bounds, and the diagonal information matrix, $\boldsymbol{\Omega}_{\text{accel}}$, allows moderate accelerations while penalizing extreme values.
Both residual terms are implemented as Ceres cost functions with automatic differentiation, allowing the optimization to jointly refine visual landmarks, IMU biases, and pose estimates while satisfying the neural network-derived motion constraints.\\

\section{Analysis and Results}
\label{sec:results}
\subsection{Dataset}
\noindent We evaluate our approach on the UZH-FPV dataset, which provides challenging high-speed drone flight sequences specifically designed to test the limits of visual-inertial SLAM systems. The dataset features aggressive maneuvers, rapid accelerations, and high-dynamic motion that are characteristic of agile UAV operations, making it particularly suitable for evaluating our neural network-based IMU integration method.
The UZH-FPV dataset contains synchronized monocular camera images and IMU measurements from a Davis event camera's integrated IMU recorded at 1000 Hz, along with ground truth trajectories obtained from a motion capture system. The sequences include various challenging scenarios such as rapid direction changes, aggressive banking maneuvers, and high-speed forward flight that cause significant motion blur and feature tracking difficulties for conventional SLAM systems.
For our experiments, we utilize the raw IMU measurements (accelerometer and gyroscope data) and ground truth trajectory information. The ground truth trajectories serve as supervision for training our neural network and as reference for quantitative evaluation of trajectory estimation accuracy.\\
We used 16 trajectories from the dataset that contain ground truth data, with a total duration of 618.97 seconds. We designate three sequences for testing: indoor forward 3 (49.5 seconds), indoor forward 6 (30.2 seconds), and indoor forward 9 (28.8 seconds).
Our experimental setup consists of two stages: network training and full system evaluation (network testing integrated with VINS-Fusion).\\
For network training, we address the limited data availability by splitting each of the 16 trajectories into three equal chunks. From the three test sequences, we reserve for test two consecutive chunks each (the first two-thirds), ensuring representation from all trajectories while increasing dataset diversity. This splitting strategy yields 546.64 seconds of training data and 72.33 seconds of test data for the network.\\
For full system evaluation, we run VINS-Fusion on the complete, unsplit versions of the three test sequences (108.6 seconds total). ICD-Net predictions are only available for the first two-thirds of each sequence-corresponding to the chunks used during network testing.
\subsection{Data Preprocessing}
\noindent \textbf{Coordinate Frame Alignment:} The UZH-FPV dataset provides ground truth in world coordinates $(W)$, while the VINS-Fusion system operates in its own coordinate frame $(SLAM)$, which is determined during its initialization process. Since the dataset does not provide IMU-to-World calibration parameters, we estimate the alignment transformation $\mathbf{R}_{W \rightarrow SLAM}$ using the available ground truth rotation data. Note that this process would not be necessary given a fixed calibration between IMU and world coordinates.\\
During the initialization phase, we collect rotation matrices $\mathbf{R}_{GT \rightarrow IMU}^k$ (from dataset ground truth) and $\mathbf{R}_{IMU \rightarrow SLAM}^k$ (from VINS-Fusion optimization) at multiple time instances. The alignment rotation is computed using Procrustes analysis by constructing the matrix:
\begin{equation}
\mathbf{M} = \sum_{k} \mathbf{R}_{IMU \rightarrow SLAM}^k \mathbf{R}_{GT \rightarrow IMU}^k
\end{equation}
We then apply SVD decomposition to $\mathbf{M}$:
\begin{equation}
\mathbf{R}_{W \rightarrow SLAM} = \mathbf{U} \mathbf{S} \mathbf{V}^T
\end{equation}
\noindent where $\mathbf{S} = \text{diag}(1, 1, \det(\mathbf{U}\mathbf{V}^T))$ ensures proper rotation.\\
\textbf{Bootstrap Phase:} Since neural network predictions represent displacements, an initial position reference is required. During the first second (first 10 samples at 10 Hz), we use SLAM position estimates as anchors:
\begin{equation}
\mathbf{p}_{NN}^i = \mathbf{p}_{SLAM}^{i-10} + \mathbf{R}_{W \rightarrow SLAM} \cdot \boldsymbol{\Delta p}_{NN}^i
\end{equation}
\noindent where $\mathbf{p}_{SLAM}^{i-10}$ is the SLAM position one second ago, $\boldsymbol{\Delta p}_{NN}^i$ is the neural network displacement prediction, and $\mathbf{R}_{W \rightarrow SLAM}$ is the alignment rotation matrix.\\
\textbf{Transition to Pure Neural Network:} After collecting 10 samples, the system transitions to using neural network positions as anchors:
\begin{equation}
\mathbf{p}_{NN}^i = \mathbf{p}_{NN}^{i-10} + \mathbf{R}_{W \rightarrow SLAM} \cdot \boldsymbol{\Delta p}_{NN}^i
\end{equation}
\noindent This maintains a continuous sliding window of neural network positions that are used later in the optimization.
\subsection{Evaluation Metrics}
\label{sec:eval_metrics}
\noindent The evaluation employs several complementary error metrics to characterize both accuracy and reliability of predictions.
\subsubsection{ICD-Net Prediction Metrics}
To assess the neural network's position prediction accuracy, we compute per-axis errors for each dimension $k \in \{x,y,z\}$. The mean absolute error (MAE) is defined as:
\begin{equation}
\text{MAE}_k = \frac{1}{N} \sum_{i=1}^{N} \big| \hat{p}_{i,k} - p_{i,k} \big|
\label{eq:mae_axis}
\end{equation}
and the median absolute error (MedAE) is:
\begin{equation}
\text{MedAE}_k = \operatorname{median}_{i=1,\dots,N} \big( \, \big| \hat{p}_{i,k} - p_{i,k} \big| \, \big)
\label{eq:medae_axis}
\end{equation}
where $\hat{p}_{i,k}$ represents the network's predicted position, $p_{i,k}$ is the ground truth position, and $N$ is the number of samples.
In addition to per-axis metrics, overall trajectory errors are assessed using the 3D Euclidean distance:
\begin{equation}
e_i = \big\| \hat{\mathbf{p}}_i - \mathbf{p}_i \big\|_2
\label{eq:euclidean_error}
\end{equation}
from which the overall MAE and overall MedAE are obtained:
\begin{align}
\text{Overall MAE} &= \frac{1}{N} \sum_{i=1}^{N} e_i \label{eq:overall_mae}\\
\text{Overall MedAE} &= \operatorname{median}_{i=1,\dots,N} (e_i) \label{eq:overall_medae}
\end{align}
\subsubsection{VINS-Fusion Trajectory Metrics}
For evaluating the complete VINS-Fusion system's trajectory estimation performance, we employ the absolute pose error (APE) metric from the evo trajectory evaluation toolkit~\cite{grupp2017evo}, a standard tool for trajectory evaluation in SLAM systems. The APE measures the Euclidean distance between the SLAM-estimated trajectory and ground truth trajectory at each timestep:
\begin{equation}
\text{APE}_i = \|\hat{\mathbf{p}}_i - \mathbf{p}_i\|_2
\label{eq:ape}
\end{equation}
where $\hat{\mathbf{p}}_i$ is the VINS-Fusion system's estimated position and $\mathbf{p}_i$ is the ground truth position at timestep $i$. The mean and maximum APE are computed as:
\begin{align}
\text{APE}_{\text{mean}} &= \frac{1}{N} \sum_{i=1}^{N} \text{APE}_i \label{eq:ape_mean}\\
\text{APE}_{\text{max}} &= \max_{i=1,...,N} \text{APE}_i \label{eq:ape_max}
\end{align}
Note that while overall MAE (\ref{eq:overall_mae}) and APE$_{\text{mean}}$ (\ref{eq:ape_mean}) are mathematically equivalent (both compute mean 3D Euclidean distance), they are applied to different system components: the former evaluates ICD-Net prediction accuracy, while the latter assesses complete VINS-Fusion trajectory performance.
\begin{figure*}[!t]
    \centering
    \includegraphics[width=\textwidth]{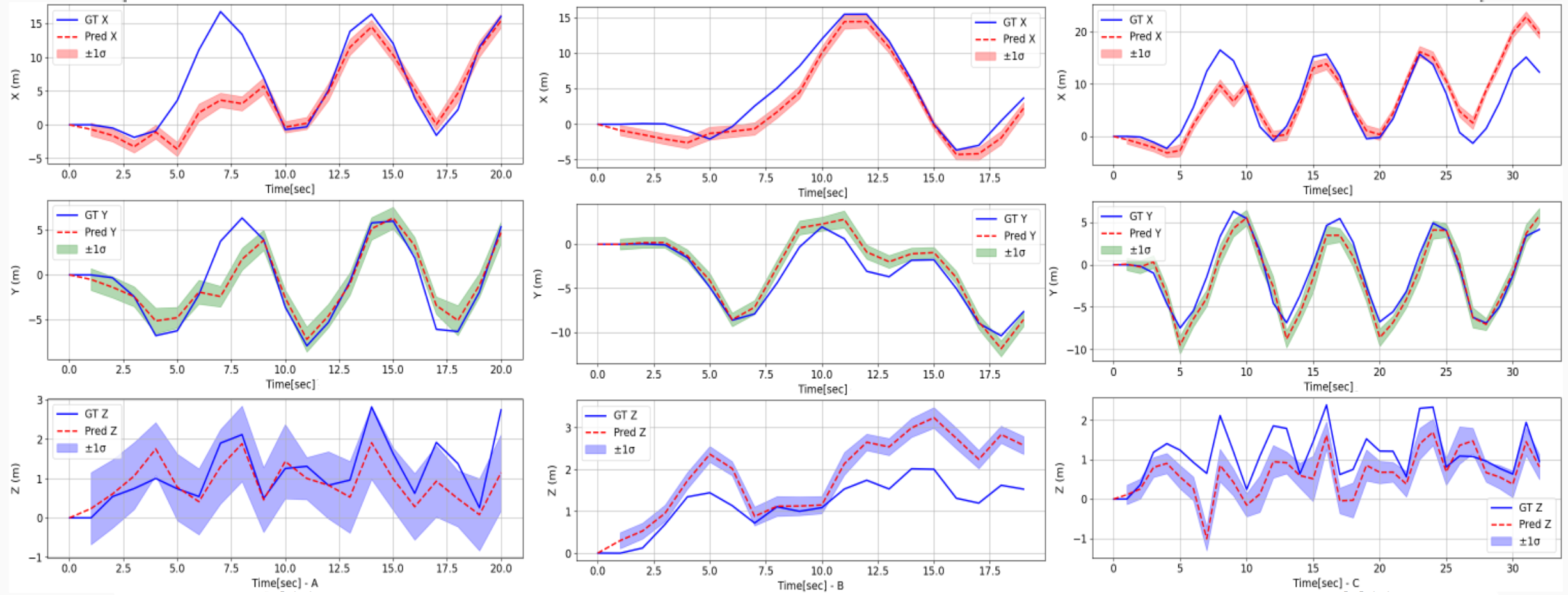}
    \caption{Comparison of ground truth (GT) and predicted (Pred) trajectories for X, Y, and Z coordinates across three different scenarios (A - indoor\_forward\_6, B - indoor\_forward\_9, and C - indoor\_forward\_3). The solid blue lines represent ground truth values, dashed red lines show predictions, and the shaded regions indicate $\pm 1\sigma$ uncertainty bounds. The predictions generally follow the ground truth trajectories with varying degrees of accuracy across the three spatial dimensions. The overall results demonstrate good performance, with the network successfully predicting the general trajectory direction across various scenarios and movement patterns.}
    \label{fig:nn_results_combined}
\end{figure*}
\begin{table*}[!t]
\renewcommand{\arraystretch}{1.3}
\caption{ICD-Net displacement prediction errors across test sequences.}
\label{tab:nn_results}
\centering
\begin{tabular}{l||ccc||ccc||cc}
\hline
\multirow{2}{*}{\textbf{Trajectory}} 
  & \multicolumn{3}{c||}{\textbf{MAE (m)}} 
  & \multicolumn{3}{c||}{\textbf{MedAE (m)}} 
  & \multicolumn{2}{c}{\textbf{Overall Metrics}} \\
\cline{2-9}
& \textbf{X} & \textbf{Y} & \textbf{Z} 
& \textbf{X} & \textbf{Y} & \textbf{Z} 
& \textbf{MAE (m)} & \textbf{MedAE (m)} \\
\hline\hline
ind\_frw\_6 & 2.90 & 1.31 & 0.44 & 1.32 & 0.74 & 0.31 & 3.47 & 1.85 \\
\hline
ind\_frw\_3 & 3.04 & 1.14 & 0.57 & 1.75 & 1.14 & 0.53 & 3.55 & 2.48 \\
\hline
ind\_frw\_9 & 1.53 & 0.95 & 0.68 & 1.18 & 0.78 & 0.89 & 2.16 & 1.99 \\
\hline\hline
\textbf{Average} & \textbf{2.49} & \textbf{1.13} & \textbf{0.56} & \textbf{1.42} & \textbf{0.89} & \textbf{0.58} & \textbf{3.06} & \textbf{2.11} \\
\hline
\end{tabular}
\end{table*}
\subsection{ICD-Net Results}
\noindent The proposed neural network architecture outputs displacement vectors and associated covariance matrices over 1-second prediction windows. For this evaluation, the network was executed at 1 Hz frequency to enable cumulative displacement integration for absolute position reconstruction. The uncertainty quantification presented as ±$1\sigma$ confidence intervals (standard deviation sleeves) represents the propagated covariance estimates from the network outputs, plotted around the computed position trajectories. Three distinct test sequences were employed to evaluate the network performance, encompassing trajectories of varying complexity and temporal characteristics. The evaluation compares predicted trajectories against ground truth data, with uncertainty bounds derived from the network's covariance predictions.\\
The evaluation employs the complementary error metrics defined in Section~\ref{sec:eval_metrics}. Table \ref{tab:nn_results} presents the quantitative performance of ICD-Net across three test sequences, demonstrating the method's capability using only inertial sensor readings. The results show that sequence ind\_frw\_9 achieves the best overall performance with an MAE of 2.16m and MedAE of 1.99m. Across all sequences, the Z-axis exhibits the strongest performance with MAE values ranging from 0.44m to 0.68m, followed by the Y-axis (0.95m to 1.31m MAE). The X-axis shows higher errors (1.53m to 3.04m MAE), with sequences ind\_frw\_3, ind\_frw\_6 presenting more challenging dynamics that result in overall MAE values of 3.47m and 3.55m respectively. The consistently lower median errors compared to mean errors across all axes indicate that the error distribution is skewed by occasional larger deviations rather than systematic bias.\\
Figure \ref{fig:nn_results_combined} illustrates the predicted trajectories derived solely from inertial sensor data, with their associated covariance sleeves compared against the ground truth along the $x$, $y$, and $z$ axes for each test sequence. The visualizations reveal that the predicted poses closely follow the ground truth references across all dimensions, successfully capturing both global trajectory structure and local motion variations. The $y$ axis demonstrates particularly tight tracking with predictions consistently within the confidence bounds across all sequences. The $z$ axis, while showing reasonable tracking of the overall trajectory pattern, exhibits less accuracy with predictions occasionally deviating from the ground truth, though they generally remain within the uncertainty sleeves. The $x$ axis shows more variation, particularly visible as lag in certain segments, though the predictions still track the overall trajectory pattern.\\
These results demonstrate that ICD-Net, utilizing only inertial sensor readings, achieves robust pose estimation capabilities under diverse motion conditions with well-calibrated uncertainty estimates. This performance is particularly significant as the method can operate as a standalone inertial odometry system or be integrated into visual-inertial frameworks such as VINS-Fusion, providing reliable pose estimates with meaningful confidence measures in scenarios where visual data may be degraded or temporarily unavailable.
\subsection{VINS-Fusion Results}
\begin{figure*}[!t]
    \centering
    \subfloat[Baseline approach.]{\includegraphics[width=0.45\textwidth]{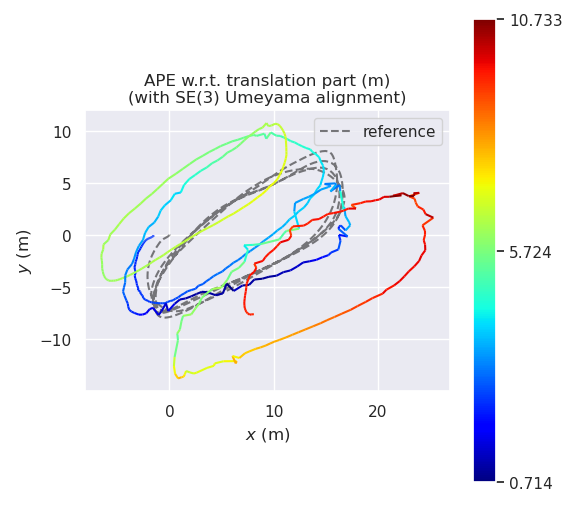}\label{fig:c_no_nn}}
    \hfil
    \subfloat[ICD-Net framework.]{\includegraphics[width=0.45\textwidth]{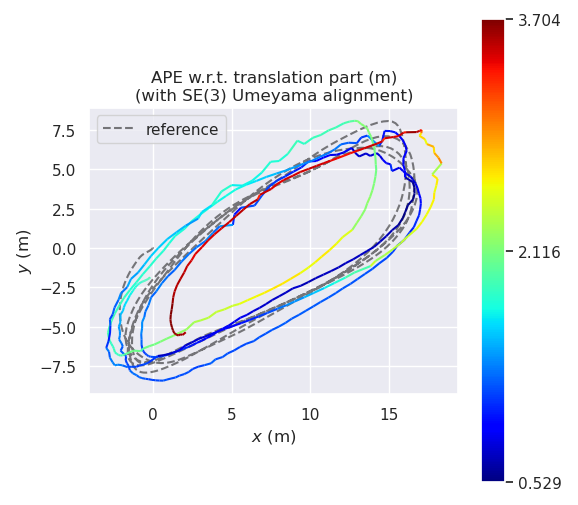}\label{fig:c_with_nn}}
    \caption{indoor\_forward\_6 trajectory comparison. The baseline is standard VINS-Fusion, while the ICD-Net framework is VINS-Fusion after integrating our network's predictions. Dashed lines represent ground truth trajectories, while colored lines show VINS-Fusion estimates. The results demonstrate significant performance improvement with reduced errors.}
    \label{fig:c_before_and_after}
\end{figure*}

\noindent We evaluated the performance of our approach by computing APE measures (\ref{eq:ape})--(\ref{eq:ape_max}) on three sequences using the evo toolkit. The evaluation compares ground truth trajectories against VINS-Fusion odometry outputs, with results summarized in Table \ref{tab:slam_results}. Our modifications demonstrate substantial improvements across all three datasets, with APE reductions ranging from 12.3\% to 67.9\%. The most dramatic enhancement is observed in indoor\_forward\_6, illustrated in Figure \ref{fig:c_before_and_after}, where both maximum and mean APE values decreased by over 65\%. These results indicate that despite imperfect network performance, our approach provides meaningful corrections that substantially enhance VINS-Fusion accuracy. The consistent improvements across diverse test scenarios validate the robustness of our method and its potential for practical visual-inertial odometry applications.

\begin{table}[!t]
\renewcommand{\arraystretch}{1.3}
\caption{VINS-Fusion APE performance evaluation results.}
\label{tab:slam_results}
\centering
\begin{tabular}{c||cc||cc||cc}
\hline
\multirow{3}{*}{\textbf{Trajectory}} & \multicolumn{2}{c||}{\textbf{Baseline}} & \multicolumn{2}{c||}{\textbf{ICD-Net (ours)}} & \multicolumn{2}{c}{\textbf{Improvement}} \\
\cline{2-7}
 & \textbf{Max} & \textbf{Mean} & \textbf{Max} & \textbf{Mean} & \textbf{Max} & \textbf{Mean} \\
 & \textbf{[m]} & \textbf{[m]} & \textbf{[m]} & \textbf{[m]} & \textbf{[\%]} & \textbf{[\%]} \\
\hline\hline
ind\_frw\_6 & 10.73 & 5.30 & 3.70 & 1.70 & 65.5\% & 67.9\% \\
\hline
ind\_frw\_3 & 8.52 & 3.11 & 3.68 & 1.98 & 56.8\% & 36.3\% \\
\hline
ind\_frw\_9 & 9.55 & 5.35 & 8.67 & 4.69 & 9.2\% & 12.3\% \\
\hline\hline
\textbf{Average} & \textbf{9.6} & \textbf{4.59} & \textbf{5.35} & \textbf{2.79} & \textbf{43.83\%} & \textbf{38.83\%} \\
\hline
\end{tabular}
\end{table}

\subsection{Camera Inconsistency}
\begin{figure*}[!t]
    \centering
    \subfloat[Baseline approach.]{\includegraphics[width=0.3\textwidth]{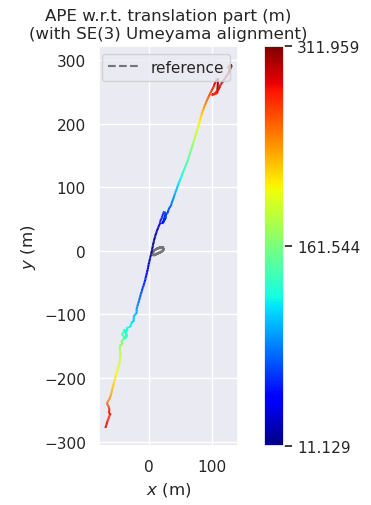}\label{fig:h_black_no_nn}}
    \hfil
    \subfloat[ICD-Net framework.]{\includegraphics[width=0.45\textwidth]{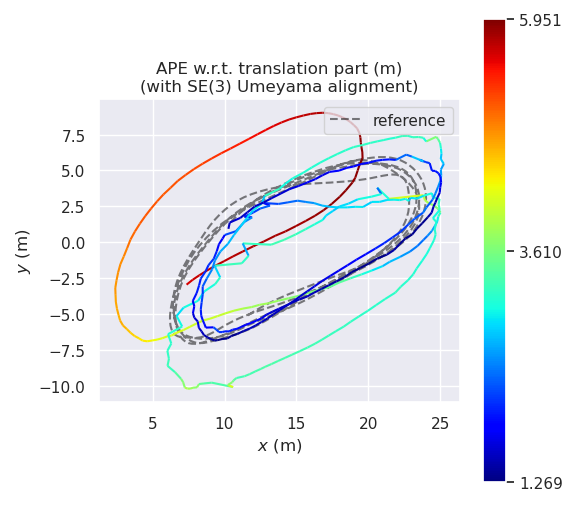}\label{fig:h_black_with_nn}}
    \caption{indoor\_forward\_3 trajectory with camera inconsistency for the (a) baseline and (b) our ICD-Net framework. The baseline VINS-Fusion produces trajectories that bear no resemblance to the ground truth circular path. The ICD-Net integrated system successfully retains circular motion patterns and remains within the ground truth operational area.}
    \label{fig:h_black_before_and_after}
\end{figure*}

\noindent To further demonstrate ICD-Net's robustness, we evaluate performance under challenging conditions by introducing periods of camera occlusion (black frames) in our test datasets. This simulates featureless environments that pose significant challenges to visual SLAM systems. During these occlusion periods, the SLAM algorithm continues to receive image data and perform optimization, generating odometry estimates without awareness that the visual information is degraded or entirely absent.
This evaluation scenario is particularly relevant as it represents real-world conditions where visual features may be temporarily unavailable due to lighting conditions, occlusions, or environmental factors. The ability to maintain tracking accuracy during such periods is crucial for robust visual-inertial odometry systems, as feature-based methods often struggle when visual information becomes unreliable or insufficient for pose estimation.
We simulated these conditions by removing between 4 to 7 seconds of data from the camera's bag file, substituting the original images with black frames. This time interval was chosen to produce measurable effects in the system's behavior. The results of VINS-Fusion integrated with our ICD-Net are summarized in Table \ref{tab:slam_black_results}. As can be seen, our approach maintains substantial improvements even under these adverse conditions, with average APE reductions of 62.7\% (maximum) and 60.6\% (mean). In particular, indoor\_forward\_3 showed exceptional resilience with an error reduction of more than 98\%. Without the ICD-Net, the system completely lost track and experienced severe drift during the visual blackout period, while with our approach the trajectory maintains consistent tracking as demonstrated in Figure \ref{fig:h_black_before_and_after}.
\begin{table}[!t]
\renewcommand{\arraystretch}{1.3}
\caption{VINS-Fusion APE performance on camera inconsistency bags.}
\label{tab:slam_black_results}
\centering
\begin{tabular}{c||cc||cc||cc}
\hline
\multirow{3}{*}{\textbf{Trajectory}} & \multicolumn{2}{c||}{\textbf{Baseline}} & \multicolumn{2}{c||}{\textbf{ICD-Net (ours)}} & \multicolumn{2}{c}{\textbf{Improvement}} \\
\cline{2-7}
 & \textbf{Max} & \textbf{Mean} & \textbf{Max} & \textbf{Mean} & \textbf{Max} & \textbf{Mean} \\
 & \textbf{[m]} & \textbf{[m]} & \textbf{[m]} & \textbf{[m]} & \textbf{[\%]} & \textbf{[\%]} \\
\hline\hline
ind\_frw\_6 & 14.95 & 9.45 & 11.43 & 6.78 & 23.5\% & 28.2\% \\
\hline
ind\_frw\_3 & 311.95 & 190.29 & 5.95 & 3.12 & 98.0\% & 98.3\% \\
\hline
ind\_frw\_9 & 42.65 & 19.19 & 14.19 & 8.57 & 66.7\% & 55.3\% \\
\hline\hline
\textbf{Average} & \textbf{123.18} & \textbf{72.98} & \textbf{10.52} & \textbf{6.16} & \textbf{62.7\%} & \textbf{60.6\%} \\
\hline
\end{tabular}
\end{table}

\section{Conclusions}
\label{sec:conc}
\noindent This paper introduced ICD-Net, a novel neural network approach that enhances visual-inertial SLAM performance by learning displacement estimates, with associated uncertainty quantification, directly from raw inertial measurements. Our method addresses fundamental limitations of analytical inertial sensor models that struggle with real-world imperfections including calibration errors, sensor noise, and rapid motion dynamics.
The key innovation lies in ICD-Net's ability to extract meaningful displacement information while simultaneously predicting measurement covariances that reflect estimation confidence. By integrating these learned constraints as additional residual terms in the VINS-Fusion optimization framework, our approach provides complementary information that compensates for various sources of error in the SLAM pipeline.\\
Beyond its integration with visual-inertial SLAM, ICD-Net demonstrates effectiveness as a standalone inertial odometry system. When operating purely on IMU measurements without visual input, the network provides displacement estimates with associated uncertainties. This inertial-only mode provides a viable fallback solution during extended visual denial periods and validates that the network genuinely learns meaningful motion representations rather than merely fitting visual-inertial correlations. Furthermore, this capability enables deployment as a standalone inertial navigation component, allowing the system to operate independently without requiring a camera in the setup.\\
Experimental validation on challenging high-speed drone sequences demonstrates substantial performance improvements over standard VINS-Fusion, with average APE reductions of 43.83\% for maximum error and 38.83\% for mean error across diverse test scenarios. Critically, our method maintains robustness under adverse conditions, as evidenced by consistent tracking performance during complete visual blackout periods where traditional approaches suffer catastrophic failure.
The uncertainty quantification component proves essential for system robustness, enabling appropriate weighting of neural network contributions relative to traditional visual and inertial terms. This demonstrates that learned displacement constraints can effectively complement existing SLAM frameworks without compromising real-time performance requirements.\\
A key strength of our ICD-Net approach lies in its versatility and broad applicability beyond the graph-based SLAM framework demonstrated in this work. The neural network's ability to provide displacement estimates and uncertainty quantification from raw IMU measurements makes it suitable for integration into various state estimation frameworks, including Kalman filter-based SLAM systems, extended Kalman filters for localization tasks, and other probabilistic estimation approaches that benefit from adaptive sensor weighting.\\
While our approach shows promising results, several limitations should be acknowledged. The ICD-Net was trained on a specific drone dataset, and the learned representations may not generalize well to other platforms with different sensor characteristics or motion dynamics. However, this limitation is partially mitigated by the relatively modest data requirements - the network achieved good performance with only 11 minutes of total training data, from a challenging dataset containing aggressive flight maneuvers, suggesting that users can feasibly collect and train on their own platform-specific datasets. Training requires ground truth trajectory data, which can be challenging to obtain without GNSS outdoors or alternative positioning systems indoors.\\
Our results validate that neural network enhancement offers a practical pathway for addressing multiple sources of SLAM degradation simultaneously. Our approach provides significant value to the SLAM community by leveraging the widely-adopted sensor fusion paradigm of camera and IMU integration, utilizing IMU data that is standard across most robotic platforms, and offering straightforward integration into existing SLAM pipelines with minimal architectural modifications. Additionally, the focus on drone applications addresses one of the most challenging domains for SLAM systems, where rapid motion, aggressive maneuvers, and dynamic environmental conditions continuously test the limits of localization accuracy. Future work will focus on extending ICD-Net and exploring its applicability to other SLAM frameworks beyond VINS-Fusion, with the ultimate goal of enabling more reliable autonomous navigation in challenging real-world environments.

\bibliographystyle{IEEEtran}
\bibliography{refs}
\end{document}